\documentclass[10pt,twocolumn,letterpaper]{article}

\usepackage[pagenumbers]{cvpr} 

\usepackage{cuted} 

\usepackage{currfile} 

\usepackage{caption} 

\usepackage{wrapfig}
\usepackage{booktabs}
\usepackage{multirow}
\usepackage{amssymb} 
\usepackage{colortbl}
\usepackage[table]{xcolor}
\usepackage{bm}
\usepackage{siunitx}
\usepackage{algorithm}
\usepackage{algpseudocode}

\usepackage[table]{xcolor}

\definecolor{rowblue}{RGB}{240,246,255}
\definecolor{rowgreen}{RGB}{253,253,253}
\definecolor{rowcream}{RGB}{255,248,232}

\definecolor{cvprblue}{rgb}{0.21,0.49,0.74}
\usepackage[pagebackref,breaklinks,colorlinks,allcolors=cvprblue]{hyperref}

\title{Spatiotemporal Message Flow  Adapter for Generalizable Face Forgery Detection}

\author{
    Kairui Shang \quad
    Xiang Wang \quad
    Jiaran Zhou \quad
    Yuezun Li\thanks{Corresponding author.}\\
    School of Computer Science and Technology, Ocean University of China \\
}

\begin{document}
\maketitle
\begin{abstract}
Adapting CLIP has emerged as a promising direction for generalizable face forgery detection. Recent methods typically design spatial adapters to identify forgery traces within individual frames or temporal adapters to detect inconsistencies across frames. As this line is still in its early stages, how to jointly model spatiotemporal cues with CLIP remains an open challenge. In this paper, we propose the Spatiotemporal Message Flow Adapter (STFlow), a framework that integrates spatial and temporal forgery evidence through bidirectional message flow between CLIP and its adapters. Our method contains three key designs: 1) a Spatial Adapter that uses two groups of learnable queries to provide precise guidance toward facial components and capture diverse forgery traces. 2) a Temporal Adapter that captures inter-frame inconsistencies by applying self-attention along the temporal dimension. 3) a Bidirectional Message Flow mechanism that allows both adapters to exchange information with CLIP and thereby enables indirect communication between adapters.
We evaluate our method on five widely used datasets and compare it with recent representative detectors. Experimental results demonstrate its effectiveness in improving generalization. We have released the code at \href{https://github.com/OUC-VAS/STFlow}{https://github.com/OUC-VAS/STFlow}.

\end{abstract}    
\section{Introduction}
\label{sec:intro}

Face forgery techniques have advanced rapidly with recent progress in generative models~\cite{goodfellow2020generative,ho2020denoising,thies2016face2face,rombach2022ldm,zhao2023diffswap}. These techniques enable adversaries to create highly realistic manipulated videos at low cost and disseminate them through social platforms, posing serious risks, including privacy violations, financial fraud, and the spread of misinformation~\cite{chesney2019deep,vaccari2020deepfakes,zhao2021multi}. Face forgery detection has therefore become an important research topic.

Recent detection methods have achieved promising results on public benchmarks through dedicated architectures and feature extraction strategies~\cite{rossler2019faceforensics++,nguyen2024laa,zhao2021multi}. However, the continual emergence of new forgery techniques presents a critical challenge: detectors often suffer substantial performance degradation on unseen forgeries, limiting their practical applicability. 
To address this challenge, many studies seek to improve generalizability by learning forensics traces shared across different forgeries~\cite{UCF,wang2023altfreezing}. Conventional solution includes crafting pseudo-fake faces~\cite{x-ray,SBI,chen2022self}, decomposition~\cite{UCF,Effort}, and frequency analysis~\cite{qian2020thinking,liu2021spatial,kim2025beyond}. With the emergence of vision foundation models such as CLIP, a more recent solution is to leverage their versatile knowledge to improve generalizability~\cite{radford2021learning,Cliping,cui2025forensics,zhu2026unleashing,DFD-FCG,Vb-StA}. 
These methods typically design spatial adapters to identify forgery traces within individual frames~\cite{cui2025forensics,zhu2026unleashing} or temporal adapters to detect inconsistencies across frames~\cite{DFD-FCG,Vb-StA}. 

\begin{figure}[!t]
    \centering
    \includegraphics[width=1\linewidth]{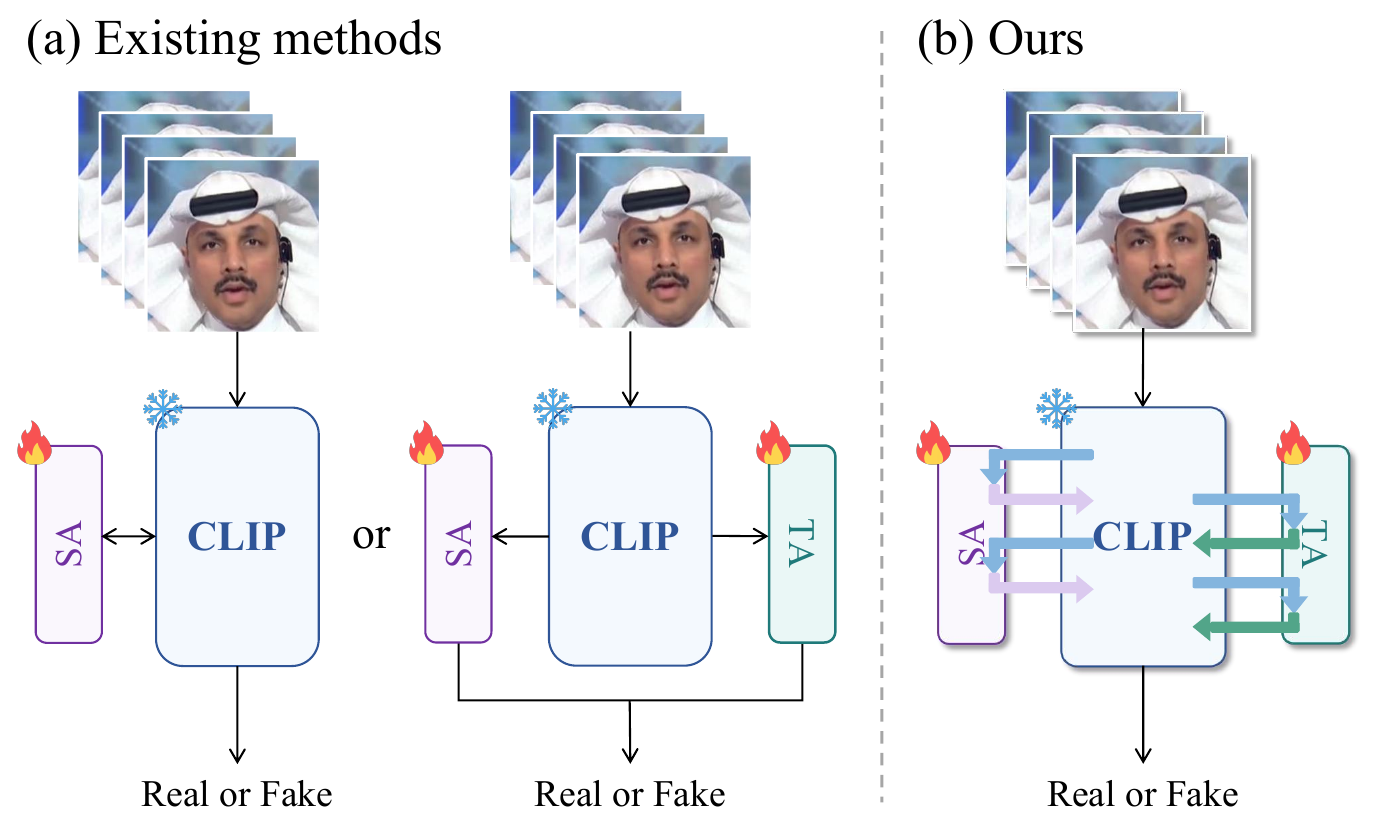}
    \vspace{-0.6cm}
    \caption{Comparison of STFlow with existing CLIP-based face forgery detectors. (a) Existing methods either use only a spatial adapter or allow limited information exchange with CLIP. (b) STFlow enables bidirectional message flow between CLIP and the spatial and temporal adapters, allowing CLIP to integrate complementary forgery cues within and across frames.}
    \label{fig:overview}
   \vspace{-0.3cm}
\end{figure}

Despite this progress, two aspects remain largely unexplored: \textbf{1) Limited spatiotemporal modeling.} 
Existing methods typically guide adapters through global objectives~\cite{Cliping}, learnable queries~\cite{DFD-FCG}, or auxiliary supervision targeting a specific forgery trace~\cite{zhu2026unleashing,x-ray}. Such guidance offers limited coverage of the diverse traces that can arise across forgery techniques, leaving the potential of adapters underexplored.
\textbf{2) Limited communication with CLIP.} 
Existing methods lack sufficient information exchange between CLIP and adapters. Some methods pass features only from CLIP to the adapter~\cite{DFD-FCG}, while others restrict the interaction to the spatial adapter~\cite{cui2025forensics} (see Fig.\ref{fig:overview}). 

\begin{figure*}[t]
    \centering
    \includegraphics[width=1\linewidth]{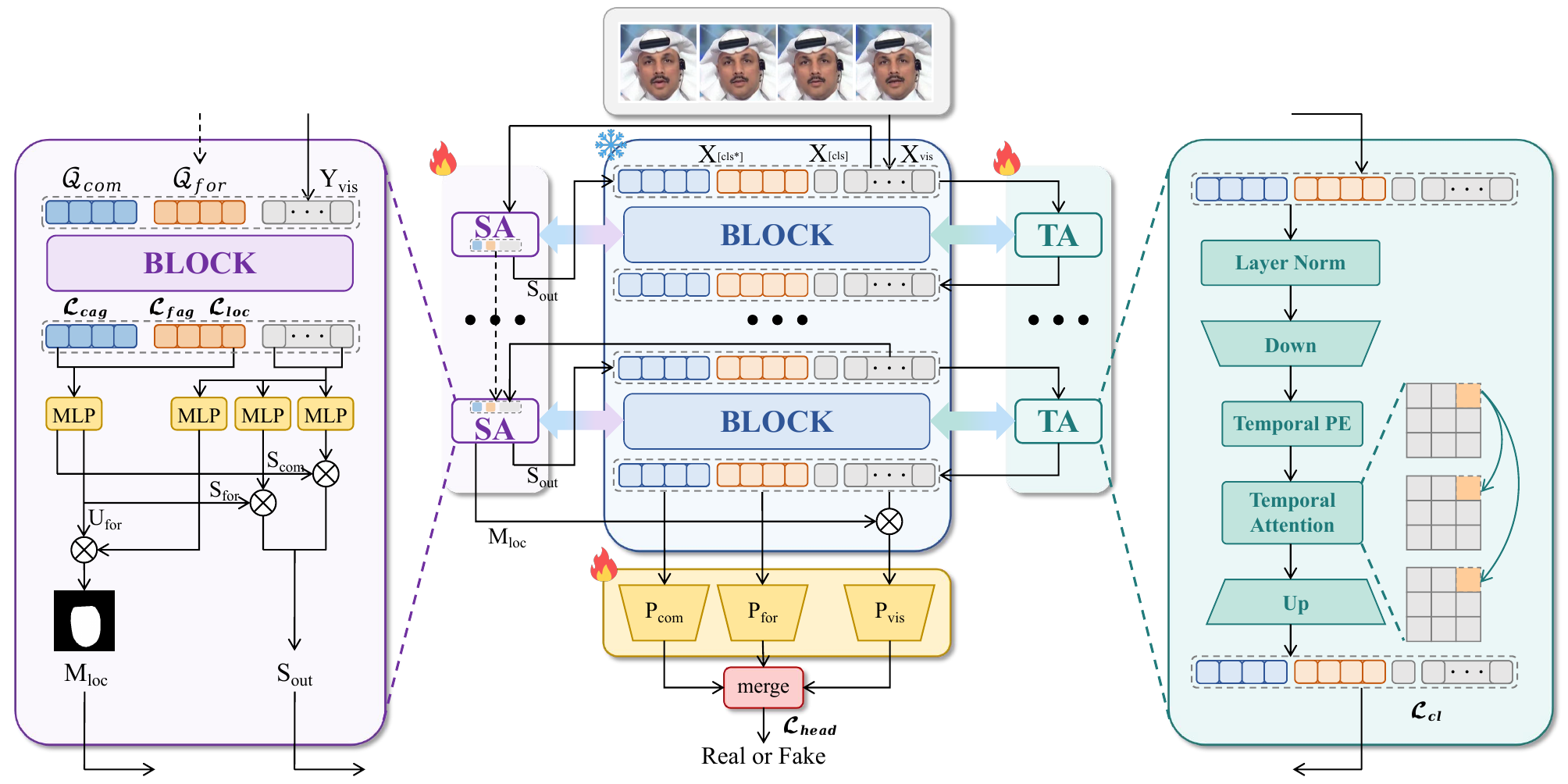}
    \vspace{-0.6cm}
    \caption{Pipeline of the proposed STFlow. The Spatial Adapter(SA) absorbs CLIP features and generates spatial guidance for CLIP. The Temporal Adapter(TA) models dependencies across frames and feeds the resulting temporal features back into CLIP. The outputs of the component, forgery, and visual heads are combined into the final prediction.}
    \label{fig:pipeline}
    \vspace{-0.3cm}
\end{figure*}

To address this gap, we propose the \textbf{S}patial-\textbf{T}emporal Message \textbf{Flow} Adapter (\textbf{STFlow}), a framework that enhances spatiotemporal evidence modeling by bidirectional message flow between adapters and CLIP (see Fig.~\ref{fig:pipeline}). Our framework has three key designs: \textbf{1) Spatial Adapter.} We use two groups of learnable queries to capture diverse spatial forgery traces. Component queries are anchored to the eye, nose, mouth, and skin regions to guide attention toward specific facial components, while
forgery queries capture diverse manipulation traces. 
\textbf{2) Temporal Adapter.} For each spatial token position, we apply self-attention across frames to model temporal dependencies and capture inter-frame inconsistencies, with contrastive learning supervision.
\textbf{3) Bidirectional
Message Flow.} Both adapters receive CLIP features and feed the spatial
and temporal cues they extract back into CLIP. This shared
pathway enables the adapters to interact indirectly through the CLIP
features they refine.

Following standard training protocols~\cite{rossler2019faceforensics++,yan2023deepfakebench,LSDA}, we train our method on FaceForensics++ (c23) and evaluate its cross-dataset generalization. We conduct experiments on five widely used datasets and compare our method with representative frame-level and video-level detection methods. The results demonstrate the effectiveness of the proposed method in improving generalizability to unseen forgeries.

\section{Related Works}
\label{sec:works}

\smallskip
\noindent\textbf{Generalizable Face Forgery Detection.} 
Face forgery detection has received considerable research attention as a means of combating manipulated media. Early methods typically relied on deep neural networks trained and evaluated on public benchmarks, using strategies such as detecting anomalous physiological signals, transforming inputs into alternative domains, and learning discriminative features directly from data~\cite{rossler2019faceforensics++,li2018ictu,qian2020thinking,afchar2018mesonet}. Although these methods perform well on established benchmarks, their performance often drops substantially when they encounter unseen forgeries. This gap limits their practical use as new forgery techniques continue to emerge.
Recent work on generalizable face forgery detection aims to close this gap by learning forensic cues shared across different forgery techniques. One common solution creates pseudo-fake faces by introducing predefined artifacts, such as blending boundaries and local inconsistencies~\cite{SBI,x-ray,haliassos2021lips}. Another uses feature decomposition to isolate shared forgery cues from training data~\cite{UCF,LSDA}. Despite this progress, reliable generalization to unseen forgeries remains an open challenge.

\smallskip
\noindent\textbf{CLIP-based Face Forgery Detection.} Adapting CLIP~\cite{radford2021learning} has emerged as a promising approach to generalizable face forgery detection. Existing methods typically keep the CLIP image encoder frozen and adapt it using parameter-efficient techniques such as adapters or low-rank decomposition~\cite{cui2025forensics, Effort, Vb-StA,houlsby2019parameter,LoRA}. These methods can be broadly grouped into frame-level and video-level solutions. 
Frame-level methods focus on spatial forgery traces within individual frames. RepDFD~\cite{RepDFD} reprograms CLIP using universal perturbations. Effort~\cite{Effort} uses an SVD-based tuning strategy to learn forgery cues while retaining CLIP's semantic knowledge. ForAda~\cite{cui2025forensics,cui26tpami} trains a lightweight side adapter using blending boundaries as its primary supervisory signal and feeds the learned forensic cues into the frozen encoder through attention. VLAForge~\cite{zhu2026unleashing} adapts CLIP to forgery detection using learnable queries and vision-language semantics.
Video-level methods also model temporal cues across frames. DFD-FCG~\cite{DFD-FCG} augments a frozen CLIP encoder with a decoder that uses four facial component queries to capture spatial cues and a patch-level temporal attention module to model cross-frame dependencies. Vb-StA~\cite{Vb-StA} uses a spatiotemporal adapter based on simple 3D convolutions.

\section{Method}
\label{sec:method}
We describe STFlow, which augments CLIP with spatiotemporal adapters and enhances their communication via a bidirectional message flow mechanism. We use ViT-L/14-336 as the visual backbone, which consists of
$H=24$ Transformer blocks.
Given a training video sample \(V=\{x^1,\dots,x^T\}\), we represent each frame $x^t$ with a classification token $X_{[\mathrm{cls}]}^t$ and a sequence of \(N\) visual patch tokens \(X^t_{\mathrm{vis}}\). To enable information exchange with the adapter, we create \(k\) copies of $X_{[\mathrm{cls}]}$ as message tokens $X^t_{[\mathrm{cls}^*]} =[{X^t_{[\mathrm{cls}]},\ldots,X^t_{[\mathrm{cls}]}}].$
The input for frame \(t\) combines all these tokens as
$X^{t,(0)} = [X^{t,(0)}_{[\mathrm{cls}^*]},\,X^{t,(0)}_{[\mathrm{cls}]},\,X^{t,(0)}_{\mathrm{vis}}]$, which contains $k+1+N$ tokens. The resulting video input is \(X^{1: T, (0)} = [X^{1,(0)},\ldots,X^{T,(0)}]\), where \((0)\) denotes the input to the first CLIP block. During the forward pass, STFlow refines the tokens, which are passed to three classification heads to predict the authenticity. 

\subsection{Spatial Adapter}
This adapter consists of four Transformer blocks and maintains two groups of $k/2$ learnable queries: component queries $Q_{\mathrm{com}}$ and forgery queries $Q_{\mathrm{for}}$. Following~\cite{DFD-FCG}, the component
queries are anchored to the eye, nose, mouth, and skin regions defined
by the 2D-FAN face parsing protocol~\cite{bulat2017far} to
provide spatial guidance. Guided by component queries, the forgery queries learn to capture
diverse manipulation traces in and around these facial regions.
Since this adapter processes each frame independently, we omit the frame index \(t\) throughout this subsection.

\smallskip
\noindent\textbf{Message Flow In.}
Before the
\(l\)-th adapter block, we fuse its visual tokens with the projected
visual tokens from the corresponding CLIP layer \(h\):
\begin{equation}
    \widetilde{Y}^{(l)}_{\mathrm{vis}}
    = Y^{(l)}_{\mathrm{vis}}
    + W^{(l)}_{\mathrm{sp}} X^{(h)}_{\mathrm{vis}},
    \label{eq:sp_clip_absorb}
\end{equation}
where \(Y^{(l)}_{\mathrm{vis}}\) denotes the visual tokens entering
block \(l\), \(Y^{(0)}_{\mathrm{vis}} = X^{(0)}_{\mathrm{vis}}\), and
\(W^{(l)}_{\mathrm{sp}}\) is a projection matrix. We concatenate the
fused visual tokens with the two query groups and pass the result
through block $l$ to obtain
$[Q^{(l+1)}_{\mathrm{com}},Q^{(l+1)}_{\mathrm{for}},
Y^{(l+1)}_{\mathrm{vis}}]$.

\smallskip
\noindent\textbf{Message Flow Out.}
After each spatial adapter block, we compute the affinity between the
updated queries and visual tokens. Omitting the block index for clarity,
we write
\begin{equation}
\begin{aligned}
    S_{\mathrm{com}}
    &= (W_q Q_{\mathrm{com}})(W_a Y_{\mathrm{vis}})^{\top},\\
    S_{\mathrm{for}}
    &= (W_q Q_{\mathrm{for}})(W_b Y_{\mathrm{vis}})^{\top},
\end{aligned}
\label{eq:scom}
\end{equation}
where $W_q$, $W_a$, and $W_b$ are learnable projections. Both affinity
matrices have shape $(k/2)\times N$, with entry $(c,n)$ giving the
affinity between query $c$ and patch $n$. We concatenate the two matrices
along the query dimension to obtain
$S_{\mathrm{out}}=[S_{\mathrm{com}};S_{\mathrm{for}}]
\in\mathbb{R}^{k\times N}$.

\begin{wrapfigure}{l}{0.4\linewidth}
 \centering
 \includegraphics[width=\linewidth]{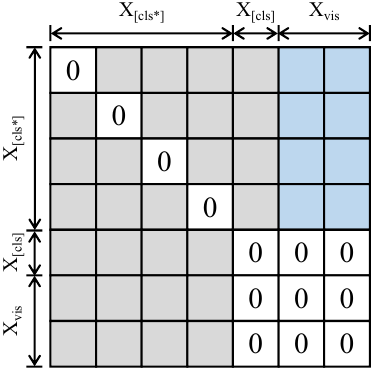}
 \vspace{-0.6cm}
    \caption{Attention bias.}
    \label{fig:attn_mask}
    \vspace{-0.3cm}
\end{wrapfigure}

We encode $S_{\mathrm{out}}$ as an additive bias
$\Delta^{(h)}\in\mathbb{R}^{(k+1+N)\times(k+1+N)}$ in the
self-attention logits of CLIP layer $h$. For each attention head,
the attention weights are $\mathrm{softmax}\left(Q^{(h)}K^{{\top}(h)} / \sqrt{d_h}+\Delta^{(h)}\right)$,
where $Q^{(h)}$ and $K^{(h)}$ are the query and key matrices and
$d_h$ is the per-head feature dimension. 
For query position $i$ and key position $j$, the value of $\Delta^{(h)}_{ij}$ is shown in Fig.~\ref{fig:attn_mask}, where gray entries denote masked ($\Delta^{(h)}_{ij}=-\infty$) connections, blue entries denote $S_{\mathrm{out}}$, and other positions are zero. The motivation is that each message token $X_{[\mathrm{cls}*]}$ attends to itself with zero added bias and to all
visual patch tokens $X_{\mathrm{vis}}$ with the corresponding biases from
$S_{\mathrm{out}}$. The classification token $X_{[\mathrm{cls}]}$ and visual patch tokens $X_{\mathrm{vis}}$ attend only to the original CLIP tokens, with zero added bias. This design injects spatial forgery cues into the message tokens while preserving
the self-attention computation among the original CLIP tokens in
this layer.

The bias guides attention from the message tokens to visual patches.

\smallskip
\noindent\textbf{Attention Supervision.}
We supervise the component and forgery affinities with separate
objectives. For each component query, we construct a target distribution
over patches from its assigned facial region. If the region covers $P$
patches, each covered patch receives weight $1/P$ and all other patches
receive zero weight. If the region covers no patches, the corresponding
target row is set to zero. Stacking the target rows gives
$\mathcal{A}\in\mathbb{R}^{(k/2)\times N}$. We define the component
attention guidance loss as
\begin{equation}
    \mathcal{L}_{\mathrm{cag}}
    =-\sum_{c,n}\mathcal{A}(c,n)
    \log\bigl[\operatorname{softmax}
    (S_{\mathrm{com}}(c,:))\bigr]_n,
\end{equation}
where the softmax is computed across patches for each query.

For forgery queries, we construct a target matrix
$\mathcal{B}\in\mathbb{R}^{(k/2)\times N}$ from the ground-truth
manipulation mask. If the mask covers $P$ patches, each covered patch
receives weight $1/P$ and all other patches receive zero weight. To
encourage different queries to learn distinct attention patterns, we
penalize the absolute cosine similarity between their attention vectors.
The forgery attention guidance loss is
\begin{equation}
\begin{aligned}
    \mathcal{L}_{\mathrm{fag}}
    ={}&-\sum_{c,n}\mathcal{B}(c,n)
    \log\bigl[\operatorname{softmax}
    (S_{\mathrm{for}}(c,:))\bigr]_n\\
    &+\sum_{c\neq c'}
    \bigl|\operatorname{cos}
    (S_{\mathrm{for}}(c,:),S_{\mathrm{for}}(c',:))\bigr|.
\end{aligned}
\end{equation}

\subsection{Temporal Adapter}
This adapter models inter-frame dependencies and feeds temporal
information back into CLIP. We insert temporal adapter blocks at selected
depths of the CLIP visual encoder. Each block projects the input tokens to a lower-dimensional space, adds learnable temporal positional embeddings, applies self-attention across frames, and projects the resulting features back to their original dimension.

\smallskip
\noindent\textbf{Message Flow In.}
At CLIP layer $h$, temporal adapter block $g$ receives the tokens from
all $T$ frames:
\begin{equation}
    [Z^{1,(g)},\ldots,Z^{T,(g)}]
    = [X^{1,(h)},\ldots,X^{T,(h)}].
\end{equation}
We first project the tokens to a lower-dimensional space and add
learnable temporal positional embeddings. We then apply self-attention
across frames at each token position. Finally, we project the resulting
tokens back to the original feature dimension to obtain
$[Z^{1,(g+1)},\ldots,Z^{T,(g+1)}]$.

\smallskip
\noindent\textbf{Message Flow Out.}
We fuse each adapter output with the corresponding CLIP token from
the same frame and token type. Omitting the frame and token indices,
we write
\begin{equation}
    X^{(h+1)}
    = X^{(h)} + \alpha
      \frac{\lVert X^{(h)}\rVert_2}
           {\lVert Z^{(g+1)}\rVert_2}
      Z^{(g+1)},
\end{equation}
where $\alpha$ is the fusion weight. The ratio of feature norms scales
the temporal update relative to the corresponding CLIP feature.

\smallskip
\noindent\textbf{Contrastive Supervision.}
To encourage the temporal adapters to learn discriminative video
representations, we map the output of each adapter to a low-dimensional
video embedding and apply a contrastive objective. Let $z_i$ denote
the embedding of video $i$, $\mathcal{R}$ the set of indices of real
videos in the current batch, and $\mathcal{V}$ the set of indices of
all valid keys. Using real videos as anchors, we define the loss as
\begin{equation}
    \mathcal{L}_{\mathrm{cl}}
    = -\sum_{i\in\mathcal{R}}\log
    \frac{\sum_{j\in\mathcal{R}}
        \exp(z_i^\top z_j/\tau_c)}
         {\sum_{j\in\mathcal{V}}
        \exp(z_i^\top z_j/\tau_c)},
\end{equation}
where $\tau_c$ is the temperature. 
This encourages the adapters to capture temporal patterns shared by
authentic videos.

\subsection{Prediction Heads}
We use three heads to predict video authenticity from the final CLIP
layer. The component head receives the first $k/2$ message tokens, the
forgery head receives the remaining $k/2$ message tokens, and the visual
head receives the visual patch tokens. For the component and forgery
heads, we apply temporal attention to the corresponding token sequences
and average the outputs over time to obtain $\widetilde{X}_{\mathrm{com}}$
and $\widetilde{X}_{\mathrm{for}}$. The two heads then use MLPs to produce
predictions $\mathcal{P}_{\mathrm{com}}$ and $\mathcal{P}_{\mathrm{for}}$.

\smallskip
\noindent\textbf{Forgery Localization for the Visual Head.}
The visual head uses a forgery localization map derived from the final
spatial adapter block. For each frame, we compute an additional affinity
matrix between forgery queries and visual tokens:
\begin{equation}
    U_{\mathrm{for}}
    = (W_q Q_{\mathrm{for}})(W_c Y_{\mathrm{vis}})^{\top},
    \label{eq:forgery_localization_affinity}
\end{equation}
where $W_c$ is a learnable projection. We reshape $U_{\mathrm{for}}$
to $(k/2)\times\sqrt{N}\times\sqrt{N}$, apply a $1\times1$ convolution
to each query map independently, and use a lightweight convolutional
head to predict a forgery map $M_{\mathrm{loc}}$. We supervise this map
with a forgery localization loss $\mathcal{L}_{\mathrm{loc}}$ that
combines binary cross-entropy~\cite{ronneberger2015u} and Dice loss~\cite{milletari2016v}.

We use $M_{\mathrm{loc}}$ to reweight the CLIP visual tokens at the
corresponding frame. The visual head then uses an MLP to produce
$\mathcal{P}_{\mathrm{vis}}$. The final video-level prediction is a
weighted combination of the three head outputs:
\begin{equation}
    \mathcal{P}
    = \lambda_1\mathcal{P}_{\mathrm{com}}
    + \lambda_2\mathcal{P}_{\mathrm{for}}
    + \lambda_3\mathcal{P}_{\mathrm{vis}},
\end{equation}
where $\lambda_1$, $\lambda_2$, and $\lambda_3$ are fusion weights.

\smallskip
\noindent\textbf{Training Objective.}
We supervise all three prediction heads with cross-entropy loss,
denoted collectively by $\mathcal{L}_{\mathrm{head}}$. The overall
training objective combines the classification loss with the spatial
attention, forgery localization, and temporal contrastive losses:
\begin{equation}
    \mathcal{L}_{\mathrm{total}}
    ={} \mathcal{L}_{\mathrm{head}}
      + \mathcal{L}_{\mathrm{cag}}
      + \mathcal{L}_{\mathrm{fag}}
     + \mathcal{L}_{\mathrm{loc}}
      + \mathcal{L}_{\mathrm{cl}},
\label{eq:training_objective}
\end{equation}

\section{Experiment}
\subsection{Experimental Setup}

\begin{table*}[!htbp]
\centering
\small
\begin{tabular}{l|c|c|ccccc|c} 
\toprule 
Method  & Venue & Input Type & CDF-v2 & DFDC & DFDCP & DFD & WDF & Avg.\\
\midrule 
Recce~\cite{Recce} & CVPR'22 & Frame & 68.71 & 69.06 & - & - &  64.31 & - \\
 \rowcolor{rowblue}  Recce~\cite{Recce} & CVPR'22 & Frame & 80.07 & 68.98 & 71.46 & 89.37 & 76.70 & 77.32 \\ 
\addlinespace[2pt]

SBI~\cite{SBI} & CVPR'22 & Frame & 93.18 & 72.42 & 86.15 & 97.56 & - &  \\
\rowcolor{rowblue}  SBI~\cite{SBI} & CVPR'22 & Frame & 79.45 & 62.70 & 76.86 & 76.35 & 73.67 & 73.81 \\
\addlinespace[2pt]

UCF~\cite{UCF} & ICCV'23 & Frame & 82.40 & 80.50 & - & 94.50 & - & - \\
\rowcolor{rowblue}  UCF~\cite{UCF} & ICCV'23 & Frame & 83.74 & 75.11 & 70.63 & 95.03 & 79.08 & 80.72 \\
\addlinespace[2pt]

LSDA~\cite{LSDA} & CVPR'24 & Frame & 91.10 & 77.00 & - & - & - & - \\
\rowcolor{rowblue}  LSDA~\cite{LSDA} & CVPR'24 & Frame & 77.96 & 71.84 & 66.46 & 84.89 & 65.64 & 73.36 \\
\addlinespace[2pt]

Effort~\cite{Effort} & ICML'25 & Frame & 95.60 & 84.30 & 90.90 & 96.50 & 84.80 & 90.42 \\
\rowcolor{rowblue}  Effort~\cite{Effort} & ICML'25 & Frame & 93.66 & 82.03 & 89.23 & 97.53 & 84.61 & 89.41 \\
\addlinespace[2pt]

RepDFD~\cite{RepDFD} & AAAI'25 & Frame & 89.94 & 80.99 & \textbf{95.03} & - & 88.05 & - \\
\rowcolor{rowblue}  RepDFD~\cite{RepDFD} & AAAI'25 & Frame & 83.81 & 78.30 & 82.64 & 91.39 & 85.24 & 84.28 \\
\addlinespace[2pt]

ForAda~\cite{cui2025forensics} & CVPR'25 & Frame & 95.70 & 87.20 & 92.90 & 97.20 & 80.50 & 90.70 \\
\rowcolor{rowblue}  ForAda~\cite{cui2025forensics} & CVPR'25 & Frame & 94.50 & 87.20 & 92.89 & 96.90 & 80.68 & 90.43 \\
\addlinespace[2pt]

 VLAForge~\cite{zhu2026unleashing} & CVPR'26 & Frame & \underline{96.80} & \underline{89.60} & - & 97.20 & - & - \\
\rowcolor{rowblue}   VLAForge~\cite{zhu2026unleashing} & CVPR'26 & Frame & 95.70 & \textbf{89.74} & 93.28 & 97.70 & 83.35 & 91.95 \\
\rowcolor{rowcream}  VLAForge~\cite{zhu2026unleashing} & CVPR'26 & Frame & 94.63 & 88.93 & 93.20 & 97.94 & 85.26 & 91.99 \\
\addlinespace[2pt]

DFD-HR~\cite{DFD-HR} & CVPR'26 & Frame & 96.00 & 86.50 & 90.10 & 98.00 & \textbf{90.70} & \underline{92.26} \\
\rowcolor{rowblue}  DFD-HR~\cite{DFD-HR} & CVPR'26 & Frame & 93.82 & 88.24 & 91.18 & 95.83 & \underline{89.61} & 91.74 \\

\midrule

TALL~\cite{TALL} & ICCV'23 & Video & 90.79 & 76.78 & - & - & - & - \\
\addlinespace[2pt]

TALL++~\cite{TALL++} & IJCV'24 & Video & 91.96 & 78.51 & - & - & - & - \\
\addlinespace[2pt]

StyleGRU~\cite{StyleGRU} & CVPR'24 & Video & 89.00 & - & - & 96.10 & - & - \\
\addlinespace[2pt]

FakeRadar~\cite{FakeRadar} & ICCV'25 & Video & 91.70 & 84.10 & 88.50 & 96.20 & - & - \\
\addlinespace[2pt]

Vb-StA~\cite{Vb-StA} & CVPR'25 & Video & 94.70 & 84.30 & 90.90 & 96.50 & 84.80 & 90.24 \\
\addlinespace[2pt]

DFD-FCG~\cite{DFD-FCG} & CVPR'25 & Video & 95.00 & 81.80 & - & - & 87.20 & - \\
\rowcolor{rowblue}  DFD-FCG~\cite{DFD-FCG} & CVPR'25 & Video & 95.20 & 80.60 & 86.60 & 94.80 & 87.40 & 88.92 \\

\midrule 

\rowcolor{rowcream} Ours (bs=8)  & - & Video & 95.34 & 84.97 & 91.63 & \textbf{98.74} & 89.12 & 91.96 \\
\rowcolor{rowcream} Ours (bs=16) & - & Video & \textbf{96.86} & 86.36 & \underline{93.34} & \underline{98.12} & 89.22 & \textbf{92.78} \\

\bottomrule 
\end{tabular}
\vspace{-0.3cm}
\caption{Cross-dataset evaluation results (video-level AUC). All methods are trained on FF++ (c23) and evaluated on five unseen datasets. The upper and lower blocks list frame-level and video-level methods. Numbers without shading are taken from the original papers, cells with a \protect\colorbox{rowblue}{blue} background report results obtained by testing the authors' released weights, and cells with a \protect\colorbox{rowcream}{yellow} background report results obtained with our own trained weights. The \textbf{best} results are indicated in bold and the \underline{second-best} results are underlined. The first and second rows of Ours correspond to batch sizes of 8 and 16.}
\label{tab:main_table} 
\vspace{-0.3cm}
\end{table*}

\smallskip
\smallskip\noindent\textbf{Datasets and Evaluation.}
Following prior work~\cite{yan2023deepfakebench,LSDA}, we use six public datasets: FaceForensics++ (FF++)~\cite{rossler2019faceforensics++}, Celeb-DF-v2 (CDF-v2)~\cite{CDF}, Deepfake Detection Challenge (DFDC)~\cite{DFDC}, DFDC Preview (DFDCP)~\cite{DFDCP}, DeepfakeDetection (DFD)~\cite{DFD}, and WildDeepfake (WDF)~\cite{WDF}. All methods are trained on the c23 version of FF++ and evaluated on the other five datasets, which contain identities and forgery methods unseen during training.
We use the area under the ROC curve (AUROC) as the primary metric. For each video, we compute a score for every sampled clip. If a video yields more than 10 clips, we discard the highest and lowest scores and average the remaining ones, otherwise, we average all clips. The resulting video-level score is used for evaluation.

\begin{table*}[t]
\setlength{\tabcolsep}{4pt} 
\small
\centering
\begin{tabular}{c|c|ccc|ccc|ccc|ccc|ccc}
\toprule
\multirow{2}{*}{SA} & \multirow{2}{*}{TA} & \multicolumn{3}{c|}{CDF-v2} & \multicolumn{3}{c|}{DFDC} & \multicolumn{3}{c|}{DFDCP} & \multicolumn{3}{c|}{DFD} & \multicolumn{3}{c}{WDF}\\ 
\cmidrule(lr){3-5} \cmidrule(lr){6-8} \cmidrule(lr){9-11} \cmidrule(lr){12-14} \cmidrule(lr){15-17}
& & AUC & AP & EER & AUC & AP & EER & AUC & AP & EER & AUC & AP & EER & AUC & AP & EER\\
\midrule
$\times$ & $\times$ & 76.54 & 87.11 & 29.46 & 74.24 & 75.94 & 32.74 & 74.80 & 86.15 & 31.05 & 87.83 & 98.48 & 19.78 & 71.37 & 78.93 & 35.40 \\  

$\checkmark$ & $\times$ & 91.15 & 94.72 & 16.81 & 78.14 & 79.53 & 29.29 & 84.03 & 91.04 & 24.61 & 96.77 & 99.62 & 8.41 & 82.39 & 87.29 & 27.41\\  

$\times$ & $\checkmark$ & 83.26 & 91.55 & 25.72 & 76.47 & 80.96 & 30.97 & 80.09 & 88.87 & 28.97 & 92.97 & 99.15 & 16.95 & 73.89 & 82.82 & 33.38\\ 

$\checkmark$ & $\checkmark$ & \textbf{95.34} & \textbf{97.64} & \textbf{11.21} & \textbf{84.97} & \textbf{87.55} & \textbf{23.89} & \textbf{91.63} & \textbf{95.10} & \textbf{18.17} & \textbf{98.74} & \textbf{99.85}& \textbf{4.65} & \textbf{89.12} & \textbf{89.73} & \textbf{17.49} \\ 

\bottomrule
\end{tabular}
\vspace{-0.3cm}
\caption{Effect of the spatial and temporal adapters. Evaluation metrics are video-level AUC, AP, and EER.}
\label{tab:STAda_table}
\vspace{-0.3cm}
\end{table*}

\begin{table*}[!t]
\small
\centering
\begin{tabular}{c | ccc | ccc | ccc} 
\toprule 
\multirow{2}{*}{SA} & \multicolumn{3}{c|}{CDF-v2} & \multicolumn{3}{c|}{DFDCP} & \multicolumn{3}{c}{DFD}\\ 
\cmidrule(lr){2-4} \cmidrule(lr){5-7} \cmidrule(lr){8-10}
& AUC & AP & EER & AUC & AP & EER & AUC & AP & EER \\
\midrule 
(w/o) com query & 93.14 & 96.28 & 14.08 & 90.31 & 94.92 & 18.79 & 96.88 & 99.63 & 9.29 \\
(w/o) ~ for query & 94.17 & 96.86 & 12.21 & 88.67 & 93.79 & 20.66 & 96.11 & 99.52 & 13.06 \\
\midrule 
2 BLOCKs & 95.30 & \textbf{97.67} & \textbf{11.21} & 90.86 & 95.02 & \textbf{18.07} & 98.03 & 99.77 & 6.84 \\
6 BLOCKs & 94.81 & 97.26 & 12.93 & 89.77 & 94.00 & 19.11 & 95.65 & 99.47 & 10.99 \\
\midrule 
Ours & \textbf{95.34} & 97.64 & \textbf{11.21} & \textbf{91.63} & \textbf{95.10} & 18.17 & \textbf{98.74} & \textbf{99.85} & \textbf{4.65} \\
\bottomrule 
\end{tabular}
\vspace{-0.3cm}
\caption{Effect of the component-query and forgery-query branches and the number of spatial adapter blocks. } 
\label{tab:SA_table}
\vspace{-0.3cm}
\end{table*}

\smallskip
\smallskip\noindent\textbf{Data Preprocessing.}
Frame-level methods typically sample a fixed number of frames per video, so the interval between adjacent frames varies with video length and no consistent temporal relation can be modeled. Instead, we sample one frame every ten frames, detect the face with RetinaFace~\cite{RetinaFacce}, and resize it to \(336\times336\). At test time, we use a sliding-window strategy that moves \(T/2\) frames per step to ensure that all intervals are covered. We apply the same set of random augmentations to the entire clip, the parameters are sampled on the first frame and replayed for the remaining \(T-1\) frames, so that the random augmentations remain consistent across the clip.

\smallskip
\smallskip\noindent\textbf{Implementation Details.}
All experiments are implemented in PyTorch 2.0.1 and run on a single RTX 3090 GPU. Each training sample is a clip of $T=8$ frames. We use the Adam optimizer with a learning rate of $3\times10^{-4}$, and train two models with batch sizes of 8 and 16. The CLIP~\cite{radford2021learning} visual encoder is ViT-L/14-336, initialized with pretrained weights and kept frozen. The CLIP layer index is denoted by $h$. The spatial and temporal adapters are trained from scratch. The spatial adapter contains four Transformer blocks. The block index is denoted by $l$, and the four blocks correspond to CLIP layers $\{6,12,18,23\}$. The temporal adapter contains six temporal attention layers. The temporal layer index is denoted by $g$, and the six layers correspond to CLIP layers $\{3,7,11,15,19,23\}$. Both adapters use an internal dimension of 256 and 8 attention heads, and we set the number of message tokens $X_{[cls*]}$ to $k=8$, including four component queries and four forgery queries. The temporal fusion weight is $\alpha=0.1$, and the contrastive temperature is $\tau_c=0.1$. The fusion weights of the three prediction heads are $\lambda_1=0.4$, $\lambda_2=0.3$, and $\lambda_3=0.3$. Two types of masks are generated for the training set only. Component localization masks \(\mathcal{A}\) are obtained by localizing the eye, nose, mouth, and skin regions with the 2D-FAN face parsing protocol and rasterizing them onto a \(24\times24\) grid. Forgery region masks \(\mathcal{B}\) are patch-level binary labels on the same grid, derived from the ground-truth manipulation mask. For a forged frame, a patch is marked as forged if more than 10\% of its pixels lie inside the manipulated region, and all patches of a real frame are marked as real.

\begin{table*}[!t]
\small
\centering
\begin{tabular}{c | ccc | ccc | ccc} 
\toprule 
\multirow{2}{*}{TA} & \multicolumn{3}{c|}{CDF-v2} & \multicolumn{3}{c|}{DFDCP} & \multicolumn{3}{c}{DFD}\\ 
\cmidrule(lr){2-4} \cmidrule(lr){5-7} \cmidrule(lr){8-10}
& AUC & AP & EER & AUC & AP & EER & AUC & AP & EER \\
\midrule 
(w/o)shallow layer & 87.72 & 93.36 & 22.27 & 88.03 & 92.89 & 20.77 & 95.01 & 99.41 & 13.62 \\
(w/o)~ ~ deep layer & 94.68 & 96.70 & \textbf{10.91} & 84.30 & 91.15 & 25.03 & 97.05 & 99.63 & 10.73 \\
\midrule 
Ours & \textbf{95.34} & \textbf{97.64} & 11.21 & \textbf{91.63} & \textbf{95.10} & \textbf{18.17} & \textbf{98.74} & \textbf{99.85} & \textbf{4.65} \\
\bottomrule 
\end{tabular}
\vspace{-0.3cm}
\caption{Effect of the temporal adapter depth. } 
\label{tab:TA_table}
\vspace{-0.3cm}
\end{table*}

\begin{table*}[ht]
\setlength{\tabcolsep}{4pt} 
\small%
\centering
\begin{tabular}{c|c|c|c|ccc|ccc|ccc}
\toprule
 \multirow{2}{*}{$\mathcal{L}_{cag}$} & \multirow{2}{*}{$\mathcal{L}_{fag}$} & \multirow{2}{*}{$\mathcal{L}_{loc}$} & \multirow{2}{*}{$\mathcal{L}_{cl}$} & \multicolumn{3}{c|}{CDF-v2} & \multicolumn{3}{c|}{DFDCP} & \multicolumn{3}{c}{DFD} \\ 
\cmidrule(lr){5-7} \cmidrule(lr){8-10} \cmidrule(lr){11-13} 
& & & & AUC & AP & EER & AUC & AP & EER & AUC & AP & EER  \\
\midrule
$\times$ & $\checkmark$& $\checkmark$ & $\checkmark$ & 94.66 & 97.25 & 13.93 & 88.90 & 93.70 & 21.29 & 97.58 & 99.71 & 9.04 \\  

$\checkmark$ & $\times$& $\checkmark$ & $\checkmark$ & 93.77 & 96.42 & 13.08 & 88.90 & 93.76 & 21.18 & 97.48 & 98.82 & 6.96 \\ 

$\checkmark$ & $\checkmark$& $\times$ & $\checkmark$ & 93.56 & 96.00 & 13.64 & 87.44 & 93.33 & 22.01 & 97.18 & 99.34 & 6.16 \\

$\checkmark$ & $\checkmark$ & $\checkmark$ & $\times$ & 92.99 & 96.23 & 14.66 & 90.26 & 94.01 & 19.42 & 97.68 & 99.72 & 10.86 \\

$\checkmark$ & $\checkmark$ & $\checkmark$ & $\checkmark$ & \textbf{95.34} & \textbf{97.64} & \textbf{11.21} & \textbf{91.63} & \textbf{95.10} & \textbf{18.17} & \textbf{98.74} & \textbf{99.85} & \textbf{4.65} \\

\bottomrule
\end{tabular}
\vspace{-0.3cm}
\caption{Effect of the four loss terms. }
\label{tab:loss}
\vspace{-0.3cm}
\end{table*}

\subsection{Compared with State-of-the-art Methods}
Table~\ref{tab:main_table} reports the cross-dataset results in video-level AUC. All methods are trained on FF++ (c23) and evaluated on five unseen datasets. The upper and lower blocks list frame-level and video-level methods, respectively. For our method, the first and second rows correspond to batch sizes of 8 and 16, and we take the batch-16 model as our main result.

Compared with frame-level methods, our method achieves an average AUC of 92.78\%, which is 0.52\% higher than the best frame-level method, DFD-HR. On CDF-v2, our batch-16 model achieves the best result (96.86\%), and on DFDCP it achieves the second-best result (93.34\%). On DFD, our batch-8 model achieves the best result (98.74\%), and the batch-16 model achieves the second-best result (98.12\%).
Compared with video-level methods, our method achieves an average AUC that is 2.54\% higher than the best video-level method, Vb-StA. Moreover, our method outperforms video-level methods on all datasets. On DFDC, our result is 2.06\% higher than Vb-StA, demonstrating robustness under severe compression. On the real-world WDF dataset, our result is 1.82\% higher than the best video-level method, DFD-FCG. These results corroborate the effectiveness and generalization ability of our method under a unified cross-dataset protocol.

\subsection{Analysis}
Due to computational resource constraints, we conduct all ablation experiments with a batch size of 8.

\smallskip\noindent\textbf{Effect of the Spatial and Temporal Adapters.}  Table~\ref{tab:STAda_table} reports the effect of the Spatial Adapter (SA) and the Temporal Adapter (TA). Using both adapters achieves the best result on all metrics across all five datasets. Relative to the baseline without either adapter, SA alone improves the average AUC and AP by 9.54\% and 5.12\%, and reduces the average EER by 8.38\%. TA alone brings smaller gains, with the average AUC and AP increasing by 4.38\% and 3.35\%, and the average EER decreasing by 2.49\%. Combining both adapters yields an average AUC of 91.96\% and an average AP of 93.97\%, improving over the baseline by 15.00\% and 8.65\%. The combined AUC and AP gains exceed the sum of the individual gains, showing that the two adapters are complementary. 

\smallskip\noindent\textbf{Effect of the Two Branches in the SA.} 
We study the component-query and forgery-query branches in Table~\ref{tab:SA_table}. Removing either branch degrades all metrics on the three datasets. Relative to the full model, removing the component queries decreases the average AUC and AP over the three datasets by 1.79\% and 0.59\%, while removing the forgery queries decreases the AUC and AP by 2.25\% and 0.81\%. The component queries are more important on CDF-v2, where removing them reduces the AUC by 2.20\%. In contrast, the forgery queries are more critical on DFDCP and DFD, where removing them reduces the AUC by 2.96\% and 2.63\%. The same dataset-dependent trend is observed for AP. These results indicate that the two branches capture complementary cues.

\smallskip\noindent\textbf{Effect of the Number of Blocks in the SA.} 
Table~\ref{tab:SA_table} compares spatial adapters with two, four, and
six Transformer blocks. The two-block adapter performs comparably to
the four-block model on CDF-v2 and slightly worse on DFDCP and DFD,
with a reduction of 0.51\% in mean AUC across the
three datasets. Increasing the number of blocks from four to six
reduces mean AUC and AP by 1.83\% and 0.62\%,
respectively. The largest AUC drop occurs on DFD, from 98.74\% to
95.65\%. The four-block configuration achieves the best overall
cross-dataset performance among the depths tested.

\smallskip\noindent\textbf{Effect of the Temporal Adapter Depth.}
Table~\ref{tab:TA_table} examines the contributions of temporal
attention blocks inserted after early CLIP layers $\{3,7,11\}$ and
later layers $\{15,19,23\}$. Removing either group reduces mean AUC
and AP across the three datasets. Relative to the full model, removing
the early blocks decreases mean AUC and AP by 4.98\% and 2.31\%, respectively, while removing the later blocks decreases them
by 3.23\% and 1.70\%. Although the early blocks contribute
more on average, the impact of each group varies across datasets.
Removing the early blocks causes the largest AUC drop on CDF-v2
(7.62\%), whereas removing the later blocks causes
the largest drop on DFDCP (7.33\%). These results
suggest that temporal modeling at different CLIP depths provides
complementary benefits for cross-dataset generalization. Retaining
all six blocks yields the highest AUC and AP on all three datasets.


\smallskip\noindent\textbf{Effect of the Losses.} 
Table~\ref{tab:loss} reports AUC, AP, and EER when each auxiliary loss
is removed in turn. Using all four losses achieves the best result on three datasets. Relative to the full model, removing $\mathcal{L}_{\mathrm{loc}}$ causes the largest degradation, decreasing the average AUC and AP over the three datasets by 2.51\% and 1.31\%. Removing $\mathcal{L}_{\mathrm{fag}}$, $\mathcal{L}_{\mathrm{cl}}$, and $\mathcal{L}_{\mathrm{cag}}$ reduces the average AUC by 1.85\%, 1.59\%, and 1.52\%. In addition, removing $\mathcal{L}_{\mathrm{cl}}$ causes the largest EER increase, the average EER rises by 3.64\%, and on DFD it increases from 4.65\% to 10.86\%, confirming the benefit of direct temporal supervision.

\begin{table}[!t]
\small%
\centering
\begin{tabular}{c|ccc} 
\toprule 
Backbone   & CDF-v2  &  DFDCP &  DFD  \\
\midrule 
 BLIP ViT-L/16 \cite{blip-14}    & 81.64 & 68.62 & 75.26 \\
 CLIP ViT-B/16 \cite{radford2021learning}  & 86.97 & 84.68 & 93.61 \\
 CLIP ViT-L/14 \cite{radford2021learning}  &\textbf{95.34} & \textbf{91.63} & \textbf{98.74} \\
\bottomrule 
\end{tabular}
\vspace{-0.3cm}
\caption{Effect of the visual backbone. } 
\label{tab:backbone} 
\vspace{-0.3cm}
\end{table}

\smallskip\noindent\textbf{Effect of the Backbones.}
We study different visual backbones in Table~\ref{tab:backbone}. CLIP ViT-L/14 achieves the best performance on all three datasets. In average AUC, it outperforms CLIP ViT-B/16 and BLIP ViT-L/16 by 6.82\% and 20.06\%. Although BLIP ViT-L/16 has a larger parameter count, its average AUC is still 13.25\% lower than that of CLIP ViT-B/16, indicating that CLIP's visual features transfer better to forgery detection. Within the CLIP family, scaling from ViT-B/16 to ViT-L/14 improves the AUC on CDF-v2, DFDCP, and DFD by 8.37\%, 6.95\%, and 5.13\%, showing that a larger ViT with a finer patch size yields stronger forgery representations. We therefore adopt CLIP ViT-L/14-336 as the default backbone.

\begin{figure}[!t]
    \centering
    \includegraphics[width=1.0\linewidth]{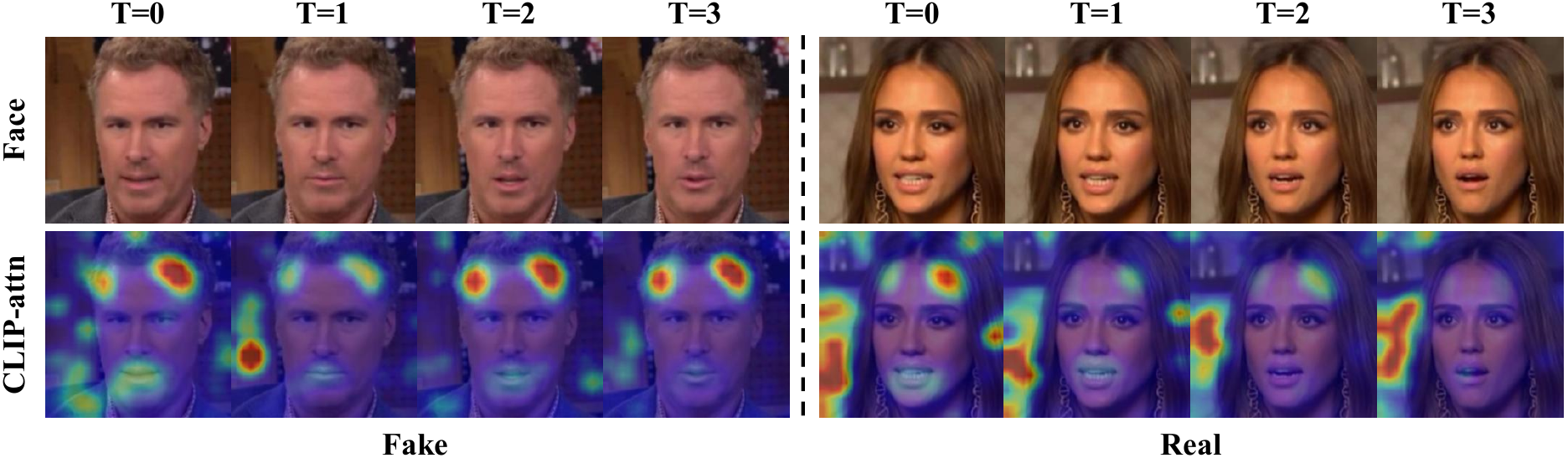}
    \vspace{-0.6cm}
    \caption{Visualization of the CLIP encoder on CDF-v2.}
    \label{fig:clip}
    \vspace{-0.3cm}
\end{figure}

\begin{figure}[!t]
    \centering
    \includegraphics[width=1\linewidth]{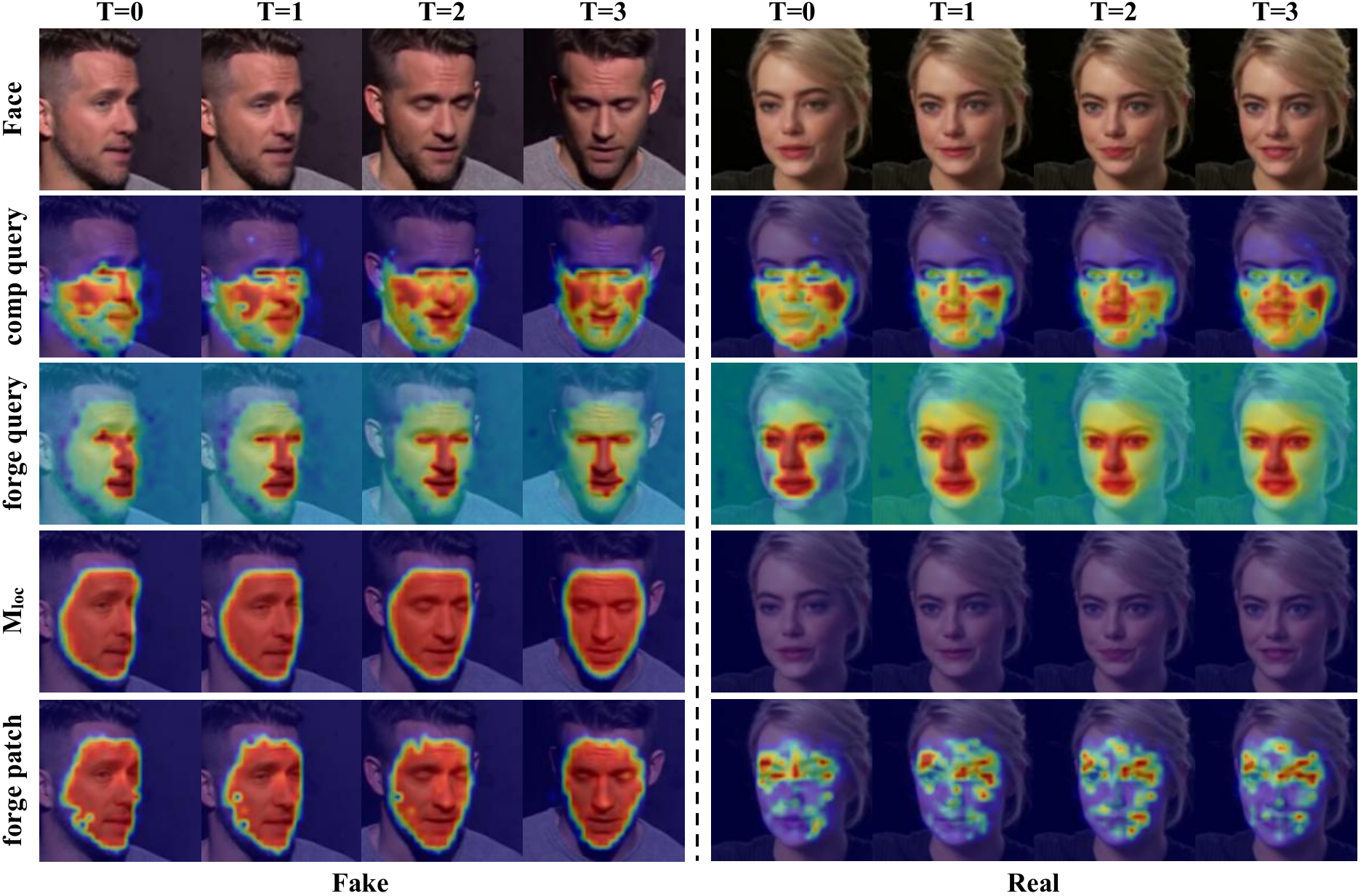}
    \vspace{-0.6cm}
    \caption{Visualization of the spatial adapter on CDF-v2.}
    \label{fig:SA}
    \vspace{-0.3cm}
\end{figure}

\smallskip\noindent\textbf{Visualization of the Spatial Adapter.}
Fig.~\ref{fig:clip} shows the attention maps of the original CLIP encoder on CDF-v2. Fig.~\ref{fig:SA} shows the attention and localization outputs of the spatial adapter on CDF-v2. Each column is a frame, and the left and right blocks show fake and real videos. The component query consistently attend to the eyes, nose, mouth, and skin regions across frames and identities, providing stable component-level guidance. The forgery query supplement the component-level guidance. The localization map covers the manipulated face region and is absent on real frames. The forge patch map yields a concentrated response over the forged face, while its responses on real faces are scattered and less coherent. The query attention patterns remain stable across frames, providing consistent spatial cues for the temporal adapter. Compared with the CLIP-only maps, the spatial adapter's attention is more stable across frames and more closely aligned with facial components and forgery traces. These results indicate that the spatial adapter turns the generic saliency of CLIP into component-aware, and forgery-relevant evidence, and that the evidence remains continuous across frames.

\begin{figure}[!t]
    \centering
    \includegraphics[width=1\linewidth]{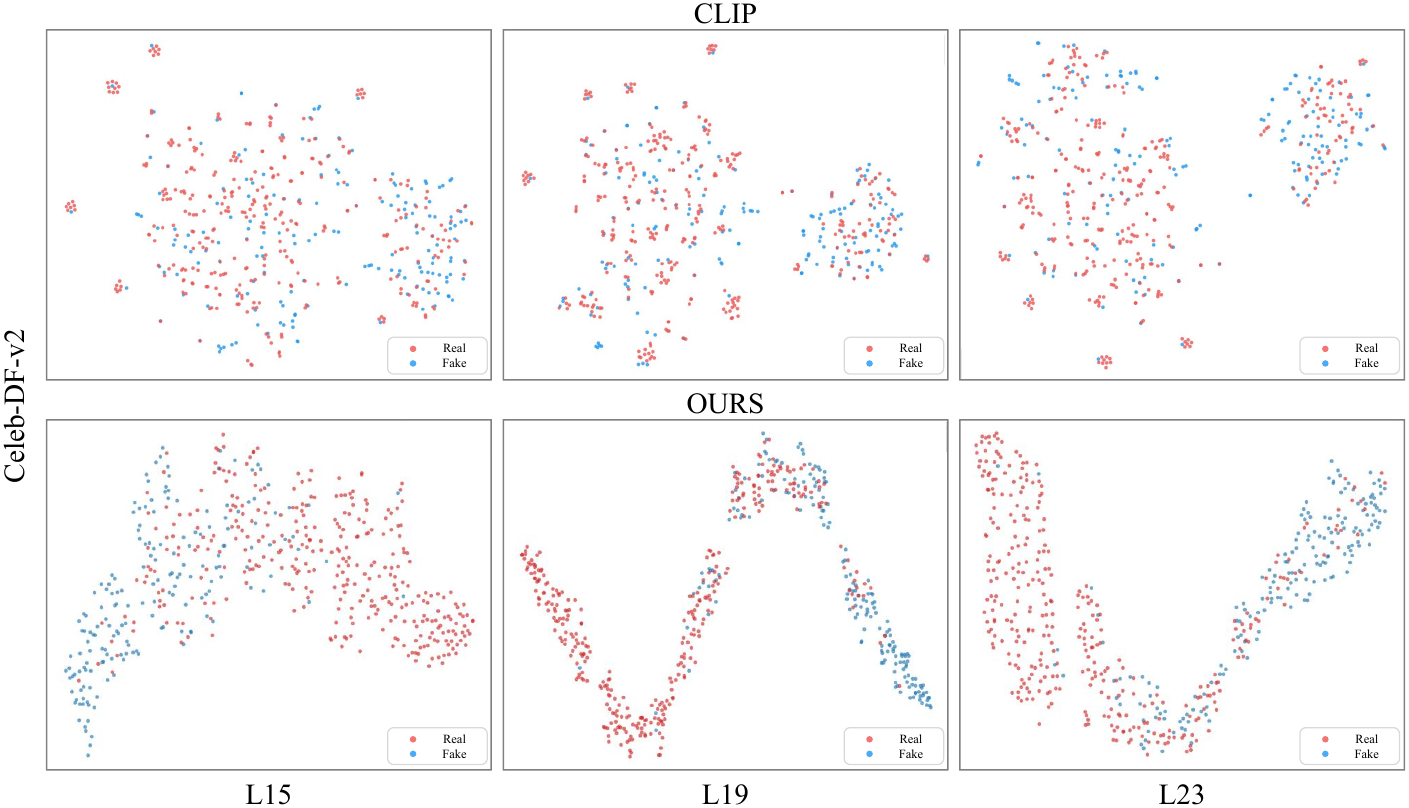}
    \vspace{-0.6cm}
    \caption{T-SNE Visualizations. The feature distribution before (top) and after (bottom) using the TA. Features are extracted from the last three layers of the TA and their corresponding layers in CLIP.}
    \label{fig:TA}
    \vspace{-0.3cm}
\end{figure}

\smallskip\noindent\textbf{T-SNE Visualization.} We use T-SNE to visualize the features of CLIP and STFlow on CDF-v2 in Fig.~\ref{fig:TA}. The top and bottom rows show the CLIP features and the features after applying the TA, layers 15, 19, and 23 are visualized. Red dots denote real videos and blue dots denote fake videos. In the CLIP features, real and fake videos are largely mixed and difficult to separate. After applying the TA, the overlap between the two classes is clearly reduced. This improvement is observed at every layer. These results show that the proposed TA makes the features more discriminative and helps the classifier distinguish unseen forgery features.

\begin{figure}[!t]
    \centering
    \includegraphics[width=1\linewidth]{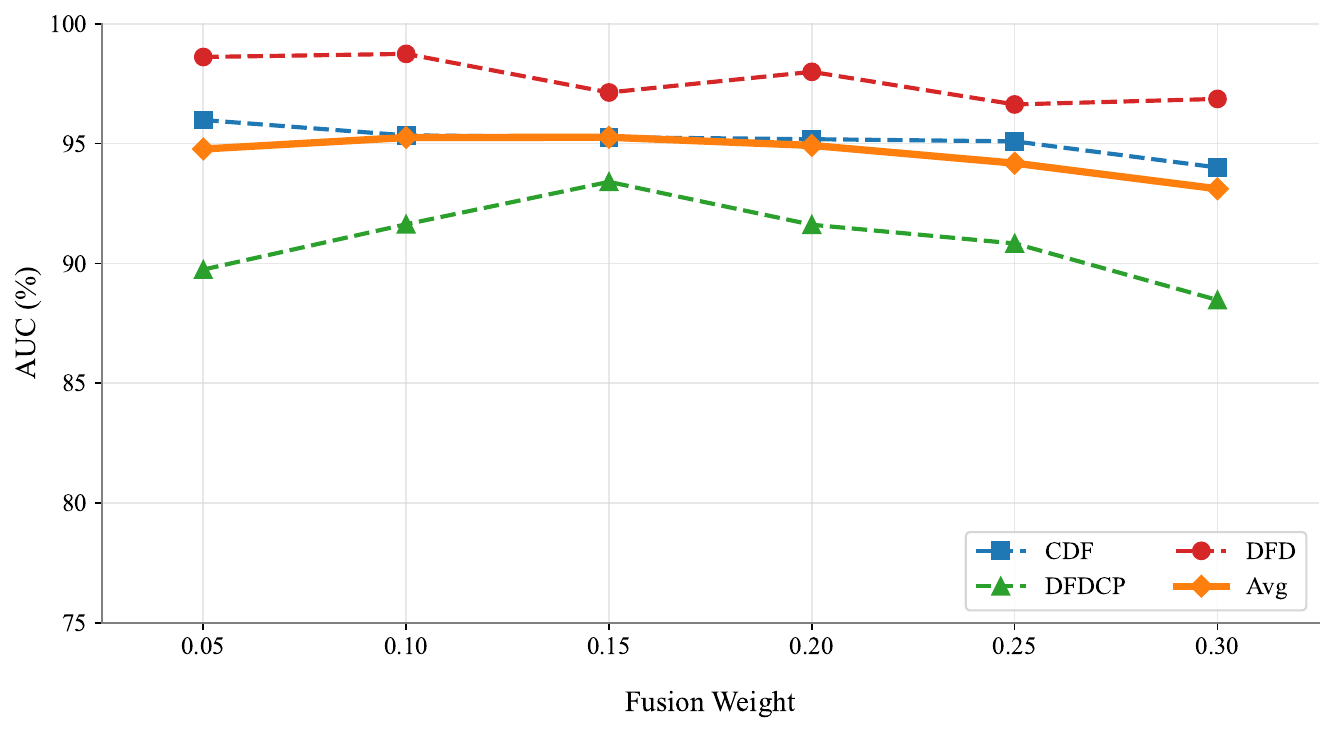}
    \vspace{-0.6cm}
    \caption{video-level AUC on CDF-v2, DFD, and DFDCP as the fusion weight varies from 0.05 to 0.30. The orange curve shows the average over the three datasets.}
    \label{fig:alpha}
    \vspace{-0.3cm}
\end{figure}

\smallskip\noindent\textbf{Effect of the Fusion Weight $\alpha$.} We study the sensitivity of the fusion weight $\alpha$ in Fig.~\ref{fig:alpha}. The average AUC remains stable at around 95\% for $\alpha\in[0.05,0.20]$, and peaks at $\alpha=0.10$--$0.15$ (95.26\%), where the results are almost identical. For $\alpha>0.20$, the average AUC decreases steadily to 93.11\%. The decline is mainly caused by DFDCP. A large $\alpha$ overweights the temporal branch and weakens the pretrained representation. We therefore set $\alpha=0.1$ in our model.

\smallskip\noindent\textbf{Limitations.}
STFlow relies on RetinaFace and 2D-FAN for facial preprocessing. Heavy occlusion and poor video quality can compromise face alignment and mask accuracy, misguiding the spatial adapter and reducing detection performance. Future work will explore an auxiliary branch that extracts forgery cues when face alignment or facial masks are unreliable.

\section{Conclusion}
We introduce the Spatiotemporal Message Flow Adapter (STFlow) for generalizable face forgery detection. Our method contains three key designs, including a spatial adapter, a temporal adapter, and a bidirectional message flow mechanism. The spatial adapter uses two groups of learnable queries to focus on specific facial components and capture diverse forgery traces. The temporal adapter applies self-attention across frames to detect temporal inconsistencies. Bidirectional message flow lets both adapters draw on CLIP features and feed their complementary forensic cues back into CLIP. Experiments on five widely used benchmarks show that STFlow generalizes better to unseen forgeries than other methods.

{
    \small
    \bibliographystyle{ieeenat_fullname}
    \bibliography{main}
}

\clearpage

\clearpage
\setcounter{page}{1}
\setcounter{section}{0}
\maketitlesupplementary

\section{Additional Results}
This section provides results on DFDC and WDF for the analysis in
Sec.~4.3 of the main paper, omitted there due to space constraints.

\smallskip\noindent\textbf{Effect of the Two Branches in the SA.} 
Table~\ref{tab:SA_table_supp} supplements the branch ablation results in Table~\ref{tab:SA_table} of the main paper. 
Removing either branch
degrades performance, with a larger average drop when the forgery-query
branch is removed. This is because component queries anchor attention to the eyes, nose, mouth, and skin, providing stable semantic guidance for the
spatial adapter. Without this guidance, attention may shift toward
irrelevant background regions. Forgery queries complement these
region-specific cues by capturing diverse traces within
and around facial components. These results align with the ones in the main paper, showing the effectiveness of the design.


\smallskip
\noindent\textbf{Effect of the Number of Blocks in the SA.}
Table~\ref{tab:SA_table_supp} also reports additional results for
spatial adapters with two, four, and six Transformer blocks.
Consistent with Table~\ref{tab:SA_table} in the main paper, the
four-block configuration offers the best balance between performance
and complexity.

\begin{table}[!ht]
\small
\centering
\setlength{\tabcolsep}{4pt} 
\resizebox{\columnwidth}{!}{ 
\begin{tabular}{c | ccc | ccc} 
\toprule 
\multirow{2}{*}{SA} & \multicolumn{3}{c|}{DFDC} & \multicolumn{3}{c}{WDF} \\ 
\cmidrule(lr){2-4} \cmidrule(lr){5-7} 
& AUC & AP & EER & AUC & AP & EER \\
\midrule 
(w/o) com query & 84.37 & 86.82 & \textbf{23.65} & 85.14 & 88.47 & 23.15 \\
(w/o) ~ for query & 81.59 & 85.06 & 26.72 & 84.69 & 89.36 & 24.49 \\
\midrule 
2 BLOCKs & \textbf{85.20} & \textbf{87.74} & 23.75 & 85.35 & 89.20 & 23.73 \\
6 BLOCKs & 84.00 & 86.99 & 23.73 & 87.80 & \textbf{90.96} & 19.79 \\
\midrule 
Ours & 84.97 & 87.55 & 23.89 & \textbf{89.12} & 89.73 & \textbf{17.49} \\
\bottomrule 
\end{tabular}
} 
\vspace{-0.3cm}
\caption{Effect of the component-query and forgery-query branches and the number of spatial adapter blocks.} 
\label{tab:SA_table_supp}
\end{table}

\smallskip
\noindent\textbf{Effect of the Temporal Adapter Depth.}
Table~\ref{tab:TA_table_supp} supplements the results in
Table~\ref{tab:TA_table} of the main paper. Across both tables,
removing the shallow temporal blocks degrades performance on CDF-v2,
DFD, and WDF, suggesting that temporal modeling of shallow layer features
helps detect subtle forgery traces. Deeper CLIP features encode
more semantic and identity-related information, so temporal attention
at these depths may capture inconsistencies such as identity drift
and facial component misalignment. Such cues may complement low-level
temporal evidence when the test domain differs substantially from
the training domain. Together, the results suggest that temporal
modeling at different depths provides complementary benefits.

\begin{table}[!t]
\small
\centering
\setlength{\tabcolsep}{4pt} 
\resizebox{\columnwidth}{!}{ 
\begin{tabular}{c | ccc | ccc} 
\toprule 
\multirow{2}{*}{TA} & \multicolumn{3}{c|}{DFDC} & \multicolumn{3}{c}{WDF} \\ 
\cmidrule(lr){2-4} \cmidrule(lr){5-7} 
& AUC & AP & EER & AUC & AP & EER \\
\midrule 
(w/o) shallow layer & 82.12 & 84.45 & 35.88 & 83.61 & 88.10 & 25.44 \\
(w/o) ~ deep layer & 81.48 & 84.73 & 26.72 & 85.70 & 88.38 & 20.68 \\
\midrule 
Ours & \textbf{84.97} & \textbf{87.55} & \textbf{23.89} & \textbf{89.12} & \textbf{89.73} & \textbf{17.49} \\
\bottomrule 
\end{tabular}
} 
\vspace{-0.3cm}
\caption{Effect of the temporal adapter depth.} 
\label{tab:TA_table_supp}
\end{table}


\smallskip
\noindent\textbf{Effect of the Losses.}
Table~\ref{tab:loss_supp} supplements the loss ablation results in
Table~\ref{tab:loss} of the main paper and shows a consistent trend,
with all losses together yielding the best AUC performance. Specifically,
$\mathcal{L}_{\mathrm{fag}}$ guides forgery queries toward ground-truth
manipulation regions, while its cosine-similarity penalty discourages
redundant attention patterns across queries. $\mathcal{L}_{\mathrm{loc}}$
supervises the localization map used to reweight visual features,
helping the classification head focus on manipulated regions.
$\mathcal{L}_{\mathrm{cag}}$ provides region-level targets that encourage
component queries to consistently attend to key facial components.
Finally, $\mathcal{L}_{\mathrm{cl}}$ directly supervises the temporal
branch, complementing the classification objective by encouraging
discriminative video representations.

\begin{table}[ht]
\setlength{\tabcolsep}{4pt} 
\small%
\centering
\resizebox{\columnwidth}{!}{ 
\begin{tabular}{c|c|c|c|ccc|ccc}
\toprule
 \multirow{2}{*}{$\mathcal{L}_{cag}$} & \multirow{2}{*}{$\mathcal{L}_{fag}$} & \multirow{2}{*}{$\mathcal{L}_{loc}$} & \multirow{2}{*}{$\mathcal{L}_{cl}$} & \multicolumn{3}{c|}{DFDC} & \multicolumn{3}{c}{WDF} \\ 
\cmidrule(lr){5-7} \cmidrule(lr){8-10} 
& & & & AUC & AP & EER & AUC & AP & EER \\
\midrule
$\times$ & $\checkmark$& $\checkmark$ & $\checkmark$ & 84.17 & 86.86 & 24.11 & 87.08 & \textbf{90.90} & 22.02 \\  

$\checkmark$ & $\times$& $\checkmark$ & $\checkmark$ & 83.17 & 85.88 & 25.76 & 86.88 & 90.62 & 22.78 \\ 

$\checkmark$ & $\checkmark$& $\times$ & $\checkmark$ & 83.22 & 85.02 & 24.51 & 86.84 & 90.80 & 21.89 \\

$\checkmark$ & $\checkmark$ & $\checkmark$ & $\times$ & 84.63 & 86.65 & 23.93 & 87.17 & 90.59 & 18.92 \\

$\checkmark$ & $\checkmark$ & $\checkmark$ & $\checkmark$ & \textbf{84.97} & \textbf{87.55} & \textbf{23.89} & \textbf{89.12} & 89.73 & \textbf{17.49} \\

\bottomrule
\end{tabular}
} 
\vspace{-0.3cm}
\caption{Effect of the four loss terms.}
\label{tab:loss_supp}
\end{table}

\smallskip\noindent\textbf{Effect of the Backbones.}
Table~\ref{tab:backbone_supp} supplements the backbone comparison in
Table~\ref{tab:backbone} of the main paper. The results follow the
same trend, with CLIP ViT-L/14 outperforming both BLIP and CLIP
ViT-B/16. Both CLIP variants also outperform BLIP. This advantage
may reflect differences in pretraining data and objectives, which
influence the transferability of visual representations.

\begin{table}[!ht]
\small%
\centering
\begin{tabular}{c|cc} 
\toprule 
Backbone   & DFDC  &  WDF  \\
\midrule 
 BLIP ViT-L/16 \cite{blip-14}    & 69.14 & 60.95 \\
 CLIP ViT-B/16 \cite{radford2021learning}  & 77.69 & 81.05 \\
 CLIP ViT-L/14 \cite{radford2021learning}  & \textbf{84.97} & \textbf{89.12} \\
\bottomrule 
\end{tabular}
\vspace{-0.3cm}
\caption{Effect of the visual backbone.} 
\label{tab:backbone_supp} 
\end{table}

\begin{figure*}[!t]
    \centering
    \includegraphics[width=1.0\linewidth]{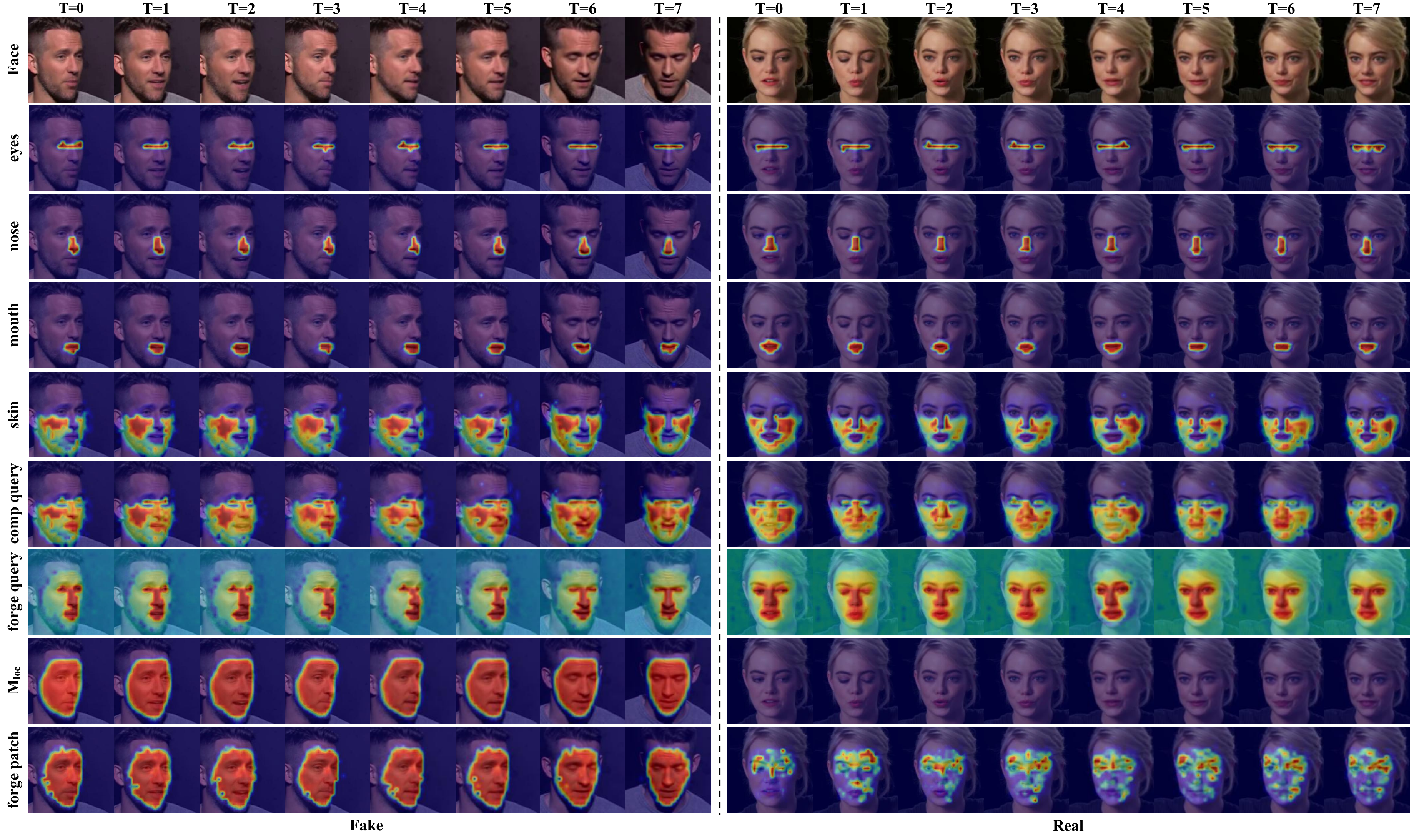}
    \vspace{-0.6cm}
    \caption{Visualization of the spatial adapter on CDF-v2.}
    \label{fig:SA_supp}
\end{figure*}

\begin{table*}[!ht]
\small
\centering
\setlength{\tabcolsep}{4pt}
\begin{tabular}{c | ccc | ccc | ccc | ccc | ccc} 
\toprule 
\multirow{2}{*}{SA} & \multicolumn{3}{c|}{CDF-v2} & \multicolumn{3}{c|}{DFDC} & \multicolumn{3}{c|}{DFDCP} & \multicolumn{3}{c|}{DFD} & \multicolumn{3}{c}{WDF}\\ 
\cmidrule(lr){2-4} \cmidrule(lr){5-7} \cmidrule(lr){8-10} \cmidrule(lr){11-13} \cmidrule(lr){14-16} 
& AUC & AP & EER & AUC & AP & EER & AUC & AP & EER & AUC & AP & EER & AUC & AP & EER\\
\midrule 
(w/o) ~ eyes  & 91.87 & 95.39 & 17.24 & 82.02 & 84.88 & 26.70 & 85.55 & 91.66 & 23.68 & 97.15 & 99.66 & 8.79 & 83.32 & 87.67 & 25.26 \\
(w/o) ~ nose  & \textbf{95.36} & 97.58 & 12.64 & 83.91 & 86.28 & 25.08 & 88.87 & 93.36 & 20.15 & 98.09 & 99.77 & 8.79 & 86.13 & \textbf{90.19} & 21.08 \\
(w/o)mouth & 94.82 & 97.39 & 13.51 & \textbf{85.56} & \textbf{87.81} & \textbf{22.65} & 90.52 & 94.66 & 19.52 & 98.21 & 99.74 & 8.16 & 86.54 & 89.65 & 19.42 \\
(w/o) ~ skin  & 93.76 & 96.60 & 15.08 & 84.72 & 87.20 & 24.87 & 91.11 & 95.03 & \textbf{16.82} & 97.54 & 99.70 & 8.66 & 84.07 & 87.78 & 23.47 \\
\midrule 
Ours & 95.34 & \textbf{97.64} & \textbf{11.21} & 84.97 & 87.55 & 23.89 & \textbf{91.63} & \textbf{95.10} & 18.17 & \textbf{98.74} & \textbf{99.85} & \textbf{4.65} & \textbf{89.12} & 89.73 & \textbf{17.49} \\
\bottomrule 
\end{tabular}
\vspace{-0.3cm}
\caption{Effect of the components. } 
\label{tab:comp_supp}
\end{table*}

\smallskip
\noindent\textbf{Visualization of the Spatial Adapter.}
Fig.~\ref{fig:SA_supp} shows the full visualization results of the spatial adapter on CDF-v2. Each column is one frame. The left and right blocks show fake and real videos. 
\textbf{Face:} the input frames. \textbf{eyes / nose / mouth / skin:} the attention maps of the four component queries. They consistently focus on the corresponding regions and remain stable across the frames. This shows that the component queries learn component level semantics rather than relying on the background. \textbf{comp query:} the aggregated attention map of the component queries. The four component queries cover the main facial components and provide a stable component level prior for the spatial branch. \textbf{forge query:} the attention map of the forgery queries. They search for diverse forgery traces and supplement the forgery traces not covered by the component queries. \textbf{$M_{\mathrm{loc}}$:} the forgery localization map. On fake videos, the response covers the forged facial region. On real videos, the response is weak. \textbf{forge patch:} the patch level forgery attention map. On fake frames, the response concentrates on the forgery facial patches. On real frames, the response is scattered.

\section{More Analysis}

\smallskip\noindent\textbf{Effect of the Component Query in Spatial Adapter.} 
This analysis examines the contribution of each facial component
in the spatial adapter. Table~\ref{tab:comp_supp} reports results
on all five datasets. Removing the eye queries causes the largest
degradation, with a mean AUC drop of 3.98\%, followed by the skin
queries with a drop of 1.72\%. Removing the nose or mouth queries
has a smaller effect, while retaining all component queries yields
the best overall performance. The contribution of eye queries may
stem from reliable landmark estimates that provide stable spatial
guidance. Skin queries cover a broader facial area, potentially
capturing more blending traces. Nose
and mouth queries cover smaller regions and may provide cues that
are partly captured by other queries. These results suggest that
facial components contribute unequally but collectively provide
complementary region-level guidance.


\smallskip
\noindent\textbf{Effect of Different Fine-tuning Strategies.}
We compare STFlow with LoRA~\cite{LoRA} and subset
fine-tuning~\cite{fine-tuning}. As shown in
Table~\ref{tab:method_supp}, both strategies achieve moderate
performance and are outperformed by STFlow on CDF-v2, DFDCP, and
DFD, where our method achieves AUCs of 95.34\%, 91.63\%, and
98.74\%, respectively. These results highlight the benefit of
explicitly modeling forgery-specific spatial and temporal cues
through our adapters while keeping the CLIP backbone frozen.

\begin{table}[!ht]
\small
\centering
\begin{tabular}{c | c | c | c} 
\toprule 
Method & CDF-v2 & DFDCP & DFD \\ 
\midrule 
LoRA~\cite{LoRA} & 86.92 & 86.73 & 91.74 \\
subset fine-tuning~\cite{fine-tuning} & 86.14 & 85.62 & 92.31 \\
\midrule 
Ours & \textbf{95.34} & \textbf{91.63} & \textbf{98.74} \\
\bottomrule 
\end{tabular}
\vspace{-0.3cm}
\caption{Effect of using various fine-tuning methods.} 
\label{tab:method_supp} 
\end{table}

\section{Algorithm Illustration}

Algorithm~\ref{alg:stflow} summarizes the training procedure of STFlow. Each frame consists of message tokens, a classification token, and visual patch tokens. These tokens are fed into the frozen CLIP and processed layer by layer. At the spatial adapter layers, \textit{Message Flow In} feeds CLIP visual features into the adapter. \textit{Message Flow Out} then injects the query generated attention bias back into CLIP. At the temporal adapter layers, \textit{Message Flow In} models temporal features across frames. \textit{Message Flow Out} fuses these temporal features back into CLIP. These two bidirectional message flows allow CLIP and the adapters to complement each other. 

\begin{algorithm}[!ht]
\caption{Training procedure of STFlow.}
\label{alg:stflow}
\begin{algorithmic}[1]
\Statex Initialize all trainable modules; freeze CLIP.
\Statex \textbf{Input:} Training clips $\{(V_i,y_i,\mathcal{A}_i,\mathcal{B}_i)\}$.
\Statex \textbf{Output:} Video-level prediction $\mathcal{P}$ and total loss $\mathcal{L}$.
\State Sample a minibatch and build tokens $[X_{[\mathrm{cls}^*]};X_{[\mathrm{cls}]};X_{\mathrm{vis}}]$ for each frame.
    \For{$h=0,\dots,H-1$}
        \If{$h\in\mathcal{D}_{\mathrm{sa}}$}
            \State \textbf{Message Flow In:} $\widetilde{Y}^{(l)}_{\mathrm{vis}}\gets Y^{(l)}_{\mathrm{vis}}+W^{(l)}_{\mathrm{sp}}X^{(h)}_{\mathrm{vis}}$.
            \State Update $Q_{\mathrm{com}},Q_{\mathrm{for}}$ with the spatial adapter block.
            \State \textbf{Message Flow Out:} compute $S_{\mathrm{out}}$ and inject bias $\Delta^{(h)}$ into CLIP attention.
        \EndIf
        \If{$h\in\mathcal{D}_{\mathrm{ta}}$}
            \State \textbf{Message Flow In:} feed CLIP tokens $X^{1:T,(h)}$ to the temporal adapter.
            \State \textbf{Message Flow Out:} fuse temporal features $Z^{1:T,(g+1)}$ into CLIP.
        \EndIf
    \EndFor
    \State $\mathcal{P}\gets\mathcal{P}_{\mathrm{com}},\mathcal{P}_{\mathrm{for}},\mathcal{P}_{\mathrm{vis}}$.
    \State $\mathcal{L}\gets\mathcal{L}_{\mathrm{head}}+\mathcal{L}_{\mathrm{cag}}+\mathcal{L}_{\mathrm{fag}}+\mathcal{L}_{\mathrm{loc}}+\mathcal{L}_{\mathrm{cl}}$.
\Statex \textbf{Return:} Video-level prediction $\mathcal{P}$.
\end{algorithmic}
\end{algorithm}

\end{document}